\documentclass{article}

\usepackage{arxiv}

\usepackage[utf8]{inputenc} 
\usepackage[T1]{fontenc}    
\usepackage{hyperref}       
\usepackage{url}            
\usepackage{booktabs}       
\usepackage{amsfonts}       
\usepackage{nicefrac}       
\usepackage{microtype}      
\usepackage{lipsum}
\usepackage{bm}
\usepackage{graphicx}%
\usepackage{multirow}%
\usepackage{amsmath,amssymb,amsfonts}%

\usepackage{amsthm}%
\usepackage{mathrsfs}%
\usepackage[title]{appendix}%
\usepackage{xcolor}%
\usepackage{textcomp}%
\usepackage{manyfoot}%
\usepackage{algorithm}%
\usepackage{algorithmicx}%
\usepackage{algpseudocode}%
\usepackage{listings}%
\usepackage{subcaption}
\usepackage{siunitx}  

\theoremstyle{thmstyleone}%

\theoremstyle{thmstyletwo}%
\theoremstyle{thmstylethree}%

\usepackage[backend=biber, style=numeric, sorting=none]{biblatex}
\graphicspath{ {./images/} }

\title{FedKDX: Federated Learning with Negative Knowledge Distillation for Enhanced Healthcare AI Systems}

\author{
 Hoang-Dieu Vu \\
  Faculty of EEE, Phenikaa School of Engineering\\
  Phenikaa University\\
  \texttt{dieu.vuhoang@phenikaa-uni.edu.vn} \\
   \And   
 Dinh-Dat Pham \\
  Faculty of EEE, Phenikaa School of Engineering\\
  Phenikaa University, Yen Nghia, Hanoi 12116, Vietnam \\
  \texttt{20010736@st.phenikaa-uni.edu.vn} \\
  \And
 Quang-Tu Pham \\
  Faculty of EEE, Phenikaa School of Engineering\\
  Phenikaa University, Yen Nghia, Hanoi 12116, Vietnam \\
  VinUni-Illinois Smart Health Center\\ 
  VinUniversity Hanoi 10000, Vietnam \\
  \texttt{tu.pq@vinuni.edu.vn} \\
  \And
 Hieu H. Pham \\
  College of Engineering \& Computer Science\\
  and VinUni-Illinois Smart Health Center\\
  VinUniversity, Hanoi 10000, Vietnam \\
  \texttt{hieu.ph@vinuni.edu.vn} \\
}

\begin{document}
\maketitle
\begin{abstract}
This paper introduces FedKDX, a federated learning framework that addresses limitations in healthcare AI through Negative Knowledge Distillation (NKD). Unlike existing approaches that focus solely on positive knowledge transfer, FedKDX captures both target and non-target information to improve model generalization in healthcare applications. The framework integrates multiple knowledge transfer techniques--including traditional knowledge distillation, contrastive learning, and NKD--within a unified architecture that maintains privacy while reducing communication costs. Through experiments on healthcare datasets (SLEEP, UCI-HAR, and PAMAP2), FedKDX demonstrates improved accuracy (up to 2.53\% over state-of-the-art methods), faster convergence, and better performance on non-IID data distributions. Theoretical analysis supports NKD's contribution to addressing statistical heterogeneity in distributed healthcare data. The approach shows promise for privacy-sensitive medical applications under regulatory frameworks like HIPAA and GDPR, offering a balanced solution between performance and practical implementation requirements in decentralized healthcare settings. The code and model are available at \url{https://github.com/phamdinhdat-ai/Fed_2024}.
\end{abstract}


\section{Introduction}

The rapid expansion of artificial intelligence in healthcare has been propelled by wearable devices that enable continuous monitoring of physiological parameters, generating substantial volumes of biomedical data for clinical decision support systems \cite{wearableAI1, wearableAI2}. However, the sensitive nature of healthcare information presents critical privacy challenges under stringent regulations such as the Health Insurance Portability and Accountability Act (HIPAA) and the General Data Protection Regulation (GDPR) \cite{HIPAA_GDPR, DataPrivacy}, which impose significant barriers to centralized data processing.

Federated Learning (FL) addresses these constraints by enabling collaborative model training across distributed devices without centralizing sensitive data \cite{FedIntroduce, FedIntroduce2}. Participating devices train models locally and share only model updates, preserving privacy while leveraging collective intelligence. Despite this promise, conventional FL approaches such as FedAvg \cite{FedAvg} suffer from high communication costs due to frequent parameter exchange and degraded performance on non-IID data distributions, which are prevalent in healthcare applications where patient profiles vary substantially \cite{FedAvg_NonIID}.

Recent advances have sought to improve FL performance in healthcare domains, yet fundamental limitations persist. FedMAT \cite{FedMAT} introduced attention-based mechanisms to learn shared and personalized features but incurs excessive communication overhead from transmitting large model parameters. FedDistill \cite{FedDistill} reduced this burden by exchanging model outputs rather than parameters and employing federated data augmentation for non-IID data, though it remains constrained by logit-based knowledge transfer that leads to information loss and slower convergence. FedKD \cite{FedKD} extended this approach by transferring knowledge through multiple channels, including hidden states and attention maps, while applying SVD-based gradient compression. However, existing distillation methods focus exclusively on positive knowledge transfer, capturing only target class information while neglecting non-target information in the model's decision boundaries. This limitation results in incomplete knowledge representation and suboptimal robustness, particularly problematic for complex patient variations and edge cases in healthcare scenarios.

To overcome these challenges, we propose FedKDX, a federated learning framework that enhances robustness and generalization in healthcare AI systems through three technical contributions. First, we introduce Negative Knowledge Distillation (NKD), a loss function that captures both target and non-target information by explicitly modeling relationships between correct and incorrect classes, providing comprehensive representation of the model's decision-making process. Second, we integrate contrastive learning mechanisms to align feature representations across distributed clients, ensuring consistent embedding spaces that facilitate knowledge transfer while preserving local personalization. Third, we employ dynamic SVD-based gradient compression to adaptively reduce communication overhead without compromising performance, making the framework practical for resource-constrained healthcare devices.

The main contributions of this work are summarized as follows:

\begin{itemize}
    \item We introduce the NKD mechanism to enable comprehensive knowledge transfer by leveraging non-target information, fundamentally addressing the blind spots of methods relying solely on positive knowledge.
    \item We establish the FedKDX unified framework, which seamlessly integrates traditional knowledge distillation, contrastive learning, and NKD, ensuring strong privacy guarantees and communication efficiency suitable for real-world deployment.
    \item Extensive experiments on three healthcare datasets demonstrate the empirical superiority of our approach, with FedKDX achieving accuracies of 94.54\%, 98.06\%, and 89.85\% on SLEEP, UCI-HAR, and PAMAP2 respectively. Notably, this represents an improvement of up to 2.53\% over state-of-the-art methods on the complex PAMAP2 dataset, while AUC scores exceeding 0.98 across all benchmarks confirm the model's high reliability.
    \item To foster reproducibility and open science, we provide the full implementation of our proposed approach, which is publicly available at \url{https://github.com/phamdinhdat-ai/Fed_2024}.
\end{itemize}

The structure of the paper is organized as follows: Section \ref{preliminaries} introduces the foundational concepts, formalizes the problem, and presents the proposed FedKDX method. Section \ref{method} elaborates on the architectural design and methodological framework in detail. Section \ref{experiment} describes the datasets employed, the preprocessing procedures, and the evaluation protocols. Section \ref{results} reports the experimental findings and provides a thorough analysis of the results. Section \ref{discussion} critically examines the strengths, limitations, and practical implications of the proposed method. Finally, Section \ref{conclusion} summarizes the key contributions of the study and outlines potential avenues for future research.

\section{Preliminaries}
\label{preliminaries}

To address the inherent challenges of statistical heterogeneity and high communication overhead in distributed learning, our work builds upon the federated learning paradigm. This section first reviews the fundamental concepts of federated learning and knowledge distillation techniques in Section \ref{fl_kd}, establishing the baseline approaches and their limitations. Subsequently, in Section \ref{problem_statement}, we articulate the specific problems that motivate our research and introduce the key principles underlying our proposed framework, FedKDX, which provides a comprehensive solution for robust and efficient decentralized training.

\subsection{Federated Learning and Knowledge Distillation}
\label{fl_kd}

Federated Averaging (FedAvg) \cite{FedAvg} establishes the foundational framework for distributed model training. In this paradigm, $K$ clients collaboratively train a shared global model with parameters $\mathbf{W}$ without exchanging raw data. Each client $k \in \mathcal{K}$ maintains a local dataset $D_k = \{(\mathbf{x}_i^{(k)}, y_i^{(k)})\}_{i=1}^{N^{(k)}}$ and minimizes its local loss function $F^{(k)}(\mathbf{W})$. At communication round $t$, the server distributes the current global model $\mathbf{W}^t$ to selected clients, which perform local training before transmitting updated parameters back for weighted aggregation. This process yields the next global model $\mathbf{W}^{t+1}$ and continues iteratively until convergence.

Despite its widespread adoption, FedAvg exhibits three fundamental limitations. First, statistical heterogeneity in non-IID data distributions across clients severely degrades convergence properties. Second, the requirement for homogeneous model architectures prevents adaptation to heterogeneous client capabilities. Third, transmitting complete model parameters each round incurs prohibitive communication costs in resource-constrained environments.

Federated Knowledge Distillation (FedKD) \cite{FedKD} addresses these challenges through a teacher-student architecture that decouples local and global models. Each client $k$ maintains a local teacher model $T_k$ with parameters $\mathbf{W}_T^{(k)}$ trained exclusively on $D_k$, while sharing a global student model $S$ with parameters $\mathbf{W}_S$. Only student model updates are communicated to the server, substantially reducing transmission overhead.

The framework employs adaptive mutual knowledge distillation where teacher and student models exchange knowledge bidirectionally. The training objective for each model comprises three components. The task loss captures standard supervised learning through cross-entropy: $L_{\text{CE},T}^{(k)}(\mathbf{W}_T^{(k)}) = \frac{1}{N^{(k)}} \sum_{i=1}^{N^{(k)}} \ell(\mathbf{W}_T^{(k)}; \mathbf{x}_i^{(k)}, y_i^{(k)})$ for teachers and $L_{\text{CE},S}^{(k)}(\mathbf{W}_S) = \frac{1}{N^{(k)}} \sum_{i=1}^{N^{(k)}} \ell(\mathbf{W}_S; \mathbf{x}_i^{(k)}, y_i^{(k)})$ for students. The distillation loss components $L_{\text{KD},T}^{(k)}$ and $L_{\text{KD},S}^{(k)}$ facilitate knowledge transfer through KL divergence over soft predictions, while hidden losses $L_{\text{Hid},T}^{(k)}$ and $L_{\text{Hid},S}^{(k)}$ align intermediate representations when architectures permit. The complete objectives are formulated as:
\begin{equation}
\begin{aligned}
L_T^{(k)} &= L_{\text{CE},T}^{(k)}(\mathbf{W}_T^{(k)}) + L_{\text{KD},T}^{(k)} + L_{\text{Hid},T}^{(k)} \\
L_S^{(k)} &= L_{\text{CE},S}^{(k)}(\mathbf{W}_S) + L_{\text{KD},S}^{(k)} + L_{\text{Hid},S}^{(k)}
\end{aligned}
\end{equation}

Communication efficiency is further enhanced through gradient-based transmission with SVD compression. The server aggregates student gradients via:
\begin{equation}
\mathbf{W}_S^{t+1} = \mathbf{W}_S^t - \eta_S \frac{1}{K} \sum_{k=1}^K \nabla L_S^{(k)}
\label{eq:ws_update}
\end{equation}
where $\eta_S$ denotes the student learning rate and $\nabla L_S^{(k)}$ represents the compressed gradient from client $k$.

FedKD demonstrates notable improvements over FedAvg through enhanced robustness to non-IID distributions via mutual distillation, support for heterogeneous architectures through independent teacher models, and reduced communication costs via selective transmission and compression. Nevertheless, two critical limitations remain that motivate our proposed approach.

\subsection{Problem Statement and Proposed Approach}
\label{problem_statement}

FedKD relies exclusively on positive knowledge transfer, aligning predictions toward correct target classes while neglecting negative knowledge from incorrect predictions and uncertainty regions. This approach overlooks valuable information about decision boundaries and class relationships. Moreover, the mutual distillation mechanism depends on soft label alignment at the output layer, which produces unstable convergence when local teachers trained on heterogeneous data generate divergent predictions. The student model receives conflicting guidance, resulting in oscillating updates and degraded performance, particularly in cross-device scenarios with severe data heterogeneity.

We propose FedKDX to address these limitations through dual stabilization mechanisms at structural and semantic levels. The framework augments FedKD's soft-label distillation with contrastive learning for structural alignment and Negative Knowledge Distillation (NKD) for semantic clarity. Contrastive learning enforces consistency in intermediate feature representations, anchoring the learning process against drift from conflicting soft labels by promoting representational agreement rather than superficial prediction matching. NKD explicitly models uncertain and incorrect predictions to sharpen decision boundaries and capture non-target class relationships. These mechanisms integrate into a unified training objective alongside traditional task loss, enabling teachers to optimize local performance while students benefit from enriched knowledge transfer across structural, semantic, and task dimensions. This approach stabilizes mutual distillation without compromising communication efficiency, providing robust decentralized training in heterogeneous and resource-constrained environments.

\section{Methodology}
\label{method}

\subsection{Overall Framework}

The FedKDX methodology enhances both communication efficiency and model performance in federated learning by integrating Knowledge Distillation, Contrastive Learning, and a novel Negative Knowledge Distillation approach, combined with dynamic gradient compression using Singular Value Decomposition (SVD).

The unified loss functions for updating the local teacher model and student model on each client are formulated as follows:
\begin{equation}
    L_T = L_{\text{CE},T} + L_{\text{KD},T} + L_{\text{NKD}} + L_{\text{CTL}}
    \label{eq:fedkdx_teacher_loss}
\end{equation}
\begin{equation}
    L_S = L_{\text{CE},S} + L_{\text{KD},S} + L_{\text{NKD}} + L_{\text{CTL}}
    \label{eq:fedkdx_student_loss}
\end{equation}

These loss functions integrate four essential components, working in concert to ensure robust knowledge transfer while balancing local learning objectives with global knowledge aggregation requirements. The cross-entropy loss $L_{\text{CE}}$ maintains standard classification performance by ensuring accurate prediction of target classes. The knowledge distillation loss $L_{\text{KD}}$ facilitates bidirectional knowledge transfer between teacher and student models. The negative knowledge distillation loss $L_{\text{NKD}}$ captures relationships between non-target classes, providing more comprehensive class understanding. The contrastive learning loss $L_{\text{CTL}}$ aligns feature representations between models to improve representational consistency.

\begin{figure*}[t]%
\centering
\includegraphics[width=0.9\textwidth]{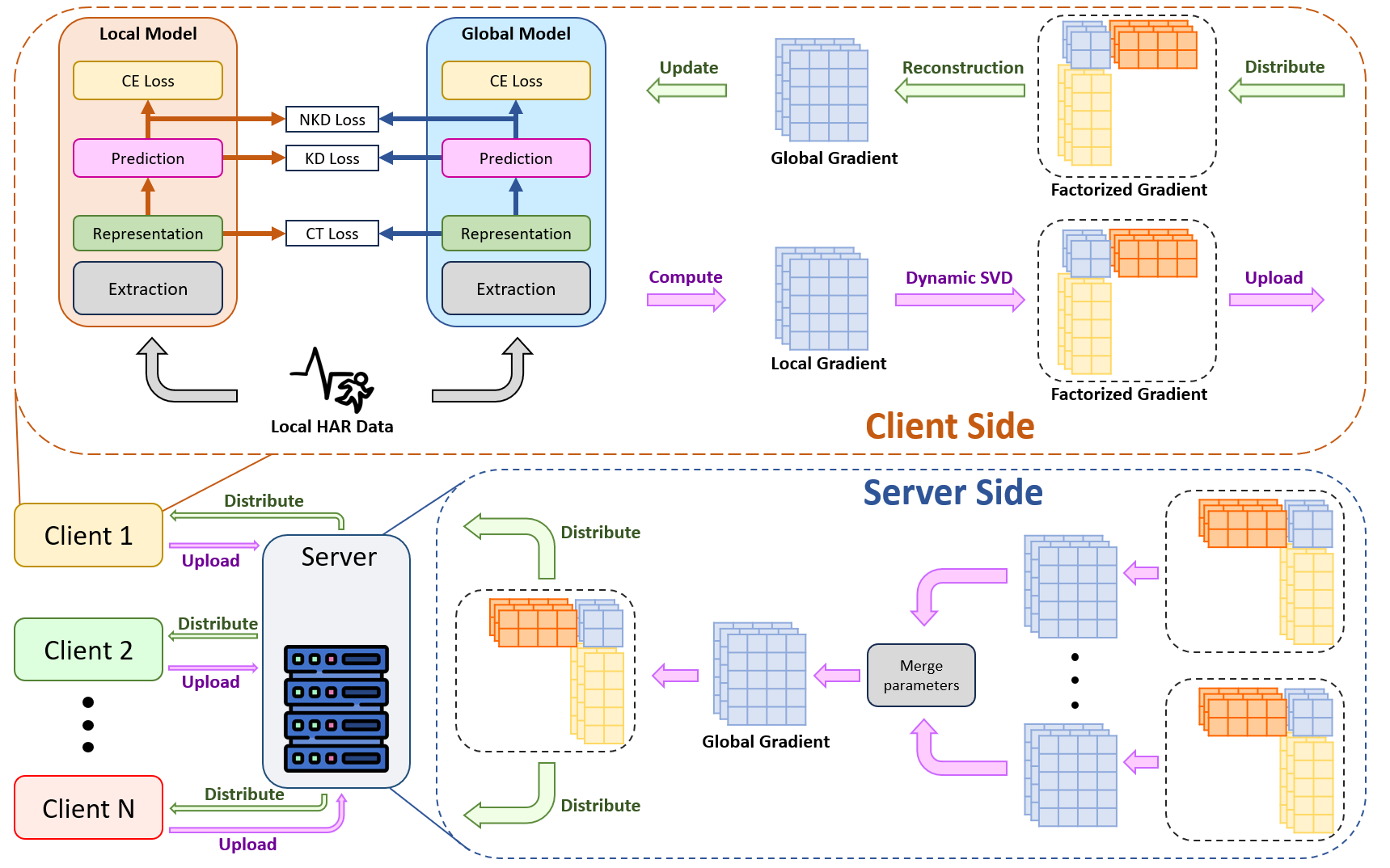}
\caption{Overview of the FedKDX workflow. This cross-device collaborative learning framework integrates knowledge distillation and gradient factorization to enable efficient federated learning for HAR. The process iterates through three phases: (1) local training with distillation from the global model, (2) gradient compression via dynamic SVD, and (3) global model refinement through aggregated updates. This cycle balances model personalization with collective learning while minimizing communication overhead.}
\label{fig:abstract}
\end{figure*}

Figure \ref{fig:abstract} provides an overview of the FedKDX framework, a cross-device collaborative learning approach that integrates federated learning, knowledge distillation, and gradient factorization for efficient Human Activity Recognition (HAR). The pipeline highlights the iterative interaction between clients and a central server. Clients train personalized models locally while compressing gradients for communication-efficient aggregation, and the server refines a shared global model to guide subsequent training rounds. This design balances model personalization with collective optimization, minimizing data exchange while preserving performance. Key components and data flows are visualized through bidirectional exchanges of factorized gradients and model updates.

The following sections provide detailed formulations for each component of the framework. Section \ref{method:knowledge_distillation} introduces the knowledge distillation setup, including the teacher-student structure and the corresponding loss formulation. Section \ref{method:contrastive_learning} presents the contrastive learning strategy and its implementation within the framework. Section \ref{method:negative_kd} describes the concept and formulation of negative knowledge distillation. Section \ref{method:gradient_compression} outlines the gradient compression technique based on singular value decomposition, including the approximation procedure. Section \ref{method:model_architecture} describes the overall model architecture and how the components are integrated. Finally, Section \ref{method:algorithm} provides the complete training algorithm, summarizing the workflow and key computational steps.

\subsection{Knowledge Distillation}
\label{method:knowledge_distillation}

Knowledge Distillation \cite{knowledge_distillation} serves as the foundational component of FedKDX, enabling bidirectional knowledge transfer between the local teacher model and the globally shared student model. This approach allows both models to benefit from each other's learned knowledge without requiring direct access to raw client data, thereby maintaining privacy while enhancing performance and generalization capabilities.

The distillation process operates on soft probability distributions generated by both models. For an input sample $\mathbf{x}_i$ on client $k$, the teacher model $T_k$ and student model $S$ produce logit vectors $\mathbf{z}_{T,i}$ and $\mathbf{z}_{S,i}$, respectively. These logits are converted to soft probability distributions using a temperature-scaled softmax function:
\begin{equation}
    \mathbf{p}_{T,i} = \operatorname{softmax}(\mathbf{z}_{T,i} / \tau)
    \label{eq:fedkdx_kdp_t}
\end{equation}
\begin{equation}
    \mathbf{p}_{S,i} = \operatorname{softmax}(\mathbf{z}_{S,i} / \tau)
    \label{eq:fedkdx_kdp_s}
\end{equation}
where $\tau$ is the temperature parameter that controls the smoothness of the probability distribution. Higher temperatures produce softer probability distributions that reveal more information about the relative confidences across classes.

The bidirectional distillation loss component $L_{\text{KD}}$ utilizes KL-Divergence to measure the discrepancy between these probability distributions in both directions \cite{knowledge_distillation}. For the teacher model, $L_{\text{KD},T}$ encourages alignment with the student's predictions, while for the student model, $L_{\text{KD},S}$ facilitates learning from the teacher's knowledge. The temperature scaling factor $\tau^2$ ensures appropriate gradient scaling during training. This bidirectional approach enables mutual learning between teacher and student models, improving both local adaptation and global generalization.

\subsection{Contrastive Learning}
\label{method:contrastive_learning}

Contrastive Learning \cite{contrastivelearning} enhances the representational capabilities of both teacher and student models by optimizing feature representations to improve similarity between corresponding features while maintaining distinctiveness across different samples. This mechanism facilitates knowledge transfer regarding data representation capabilities between the models, providing structural alignment that stabilizes the learning process in heterogeneous federated environments.

For a given input sample $\mathbf{x}_i$, the teacher and student models generate feature representations $\mathbf{h}_{T,i} = f_T(\mathbf{x}_i)$ and $\mathbf{h}_{S,i} = f_S(\mathbf{x}_i)$ through their respective feature extraction functions $f_T$ and $f_S$. The InfoNCE loss function $L_{\text{CTL}}$ minimizes the distance between positive pairs, defined as representations from the same sample by different models, while maximizing the distance between negative pairs, defined as representations from different samples.

For a batch of $B$ samples, the contrastive loss for sample $i$ is formulated as:
\begin{equation}
    L_{\text{CTL}} = -\log \frac{\exp(\operatorname{sim}(\mathbf{h}_{T,i}, \mathbf{h}_{S,i}) / \tau)}{\sum_{j=1}^{B} \exp(\operatorname{sim}(\mathbf{h}_{T,i}, \mathbf{h}_{S,j}) / \tau)}
    \label{eq:fedkdx_ctl}
\end{equation}
The similarity function $\operatorname{sim}(\cdot, \cdot)$ is typically cosine similarity, and $\tau$ is the temperature scaling parameter that controls the sharpness of the probability distribution. The denominator includes all student representations in the current batch, creating a comprehensive contrastive learning framework. By enforcing alignment between internal feature representations of teacher and student models, contrastive learning acts as a robust anchor that prevents drift from noisy soft labels, promoting fundamental representational consistency rather than superficial prediction agreement.

\subsection{Negative Knowledge Distillation}
\label{method:negative_kd}

Negative Knowledge Distillation extends traditional knowledge distillation by incorporating non-target knowledge transfer, explicitly modeling the relationships among classes that are not the ground truth. This approach addresses limitations of conventional knowledge distillation, which focuses exclusively on positive knowledge transfer toward target classes, by ensuring that the student model learns not only about the target class but also understands the complete probability distribution over all classes. This creates more comprehensive class understanding and sharper decision boundaries that better match the teacher model's learned representations.

The method operates on soft probability distributions $\mathbf{p}_{T,i}$ and $\mathbf{p}_{S,i}$ for sample $\mathbf{x}_i$, calculated using the temperature parameter $\tau$ as established in Equations (\ref{eq:fedkdx_kdp_t}) and (\ref{eq:fedkdx_kdp_s}). Given $C$ total classes and $c^*$ as the index of the true target class, the proposed $L_{\text{NKD}}$ loss is formulated as:
\begin{equation}
    L_{\text{NKD}} = -(1 - \gamma) \cdot [\mathbf{p}_{T,i}]_{c^*} \log([\mathbf{p}_{S,i}]_{c^*}) - \gamma \cdot \tau^2 \cdot \sum_{c \neq c^*}^{C} [N(\mathbf{p}_{T,i})]_c \log([N(\mathbf{p}_{S,i})]_c)
    \label{eq:fedkdx_nkd}
\end{equation}
This formulation incorporates both positive and negative knowledge transfer through a balanced approach. The first term represents positive knowledge transfer, encouraging the student to match the teacher's confidence on the target class. The second term represents negative knowledge transfer, ensuring that the student learns the teacher's understanding of relationships among non-target classes. The hyperparameter $\gamma$ controls the balance between these two types of knowledge, allowing flexible adaptation to different learning scenarios.

The normalized probability distribution $N(\mathbf{p})$ over all non-target classes is defined as $[N(\mathbf{p})]_c = [\mathbf{p}]_c / \sum_{k \neq c^*}[\mathbf{p}]_k$ for any class $c \neq c^*$, effectively redistributing the probability mass among incorrect classes to focus on their relative relationships. This normalization ensures that the negative knowledge transfer component properly captures the teacher's understanding of which incorrect classes are more or less likely, independent of the confidence assigned to the target class.

Key notation includes $[\mathbf{p}]_c$ denoting the probability of class $c$ in distribution $\mathbf{p}$, $\gamma$ representing the hyperparameter balancing positive and negative knowledge transfer, $N(\mathbf{p})$ indicating the normalized probability distribution over all non-target classes, and $\tau$ serving as the temperature parameter maintaining consistency with knowledge distillation. The mechanism of negative knowledge distillation works by explicitly pushing the student model away from incorrect classes in proportion to how confidently the teacher model rejects them, creating more refined decision boundaries and reducing semantic ambiguity in uncertain regions of the feature space. This comprehensive formulation ensures that the student model learns complete class relationships, addressing issues such as information loss and slow convergence that may occur in traditional positive knowledge-focused knowledge distillation methods.

\subsection{Gradient Compression with SVD}
\label{method:gradient_compression}

Singular Value Decomposition (SVD) \cite{SVD} is integrated into the FedKDX framework to reduce communication costs in federated learning by compressing gradients before transmission. This technique leverages the low-rank characteristics of model parameters to achieve significant size reduction while maintaining model performance and controlling approximation error within acceptable bounds.

The compression process operates on gradient matrices through low-rank factorization. For a gradient $\mathbf{G}_i$ represented as a matrix with $P$ rows and $Q$ columns (assuming $P \geq Q$), SVD decomposes it into factor matrices $\mathbf{U}_i \in \mathbb{R}^{P \times R}$, $\mathbf{\Sigma}_i \in \mathbb{R}^{R \times R}$, and $\mathbf{V}_i \in \mathbb{R}^{R \times Q}$, where $R$ is the number of retained singular values. The compression is beneficial when the condition $PR + R^2 + RQ < PQ$ is satisfied, ensuring that the compressed representation requires less storage than the original gradient matrix.

To control approximation error, an energy threshold $\epsilon$ determines the number of singular values to retain. This threshold ensures that the preserved energy ratio exceeds a minimum acceptable level:
\begin{equation}
\frac{\sum_{j=1}^{R} \sigma_j^2}{\sum_{j=1}^{Q} \sigma_j^2} > \epsilon
\label{eq:fedkdx_threshold}
\end{equation}
where $\sigma_j$ are the singular values of $\mathbf{G}_i$ arranged in descending order. This energy-based criterion provides theoretical guarantees on the approximation quality while allowing adaptive compression based on the intrinsic dimensionality of the gradient structure.

The framework implements a dynamic threshold strategy to support model convergence throughout the training process. The threshold $\epsilon(\rho)$ varies according to training progress:
\begin{equation}
\epsilon(\rho) = \epsilon_{\text{start}} + (\epsilon_{\text{end}} - \epsilon_{\text{start}}) \cdot \rho, \quad \rho \in [0, 1]
\label{eq:fedkdx_adaptiveT}
\end{equation}
where $\epsilon_{\text{start}}$ and $\epsilon_{\text{end}}$ are hyperparameters defining the initial and final threshold values, and $\rho$ represents the training progress percentage. This dynamic approach constitutes a key contribution of our work in adaptive optimization. By allowing the model to learn from coarsely approximated gradients with high compression rates early in training and progressively more accurate gradients with lower compression rates as convergence approaches, the framework achieves improved final model accuracy while maintaining communication efficiency throughout the training process. The adaptive threshold strategy effectively balances the trade-off between communication cost and model performance across different training phases, ensuring efficient resource utilization without compromising convergence quality.

\subsection{Model Architectures}
\label{method:model_architecture} 

To address human activity recognition (HAR) from multivariate time-series data, we propose a convolutional neural network (CNN) architecture optimized for automatic feature extraction from sensor data. The CNN framework was selected for its capability to efficiently learn local temporal patterns while maintaining computational efficiency—a critical requirement for federated learning environments with constrained resources and limited communication bandwidth.

\begin{figure*}[t]%
\centering
\includegraphics[width=0.9\textwidth]{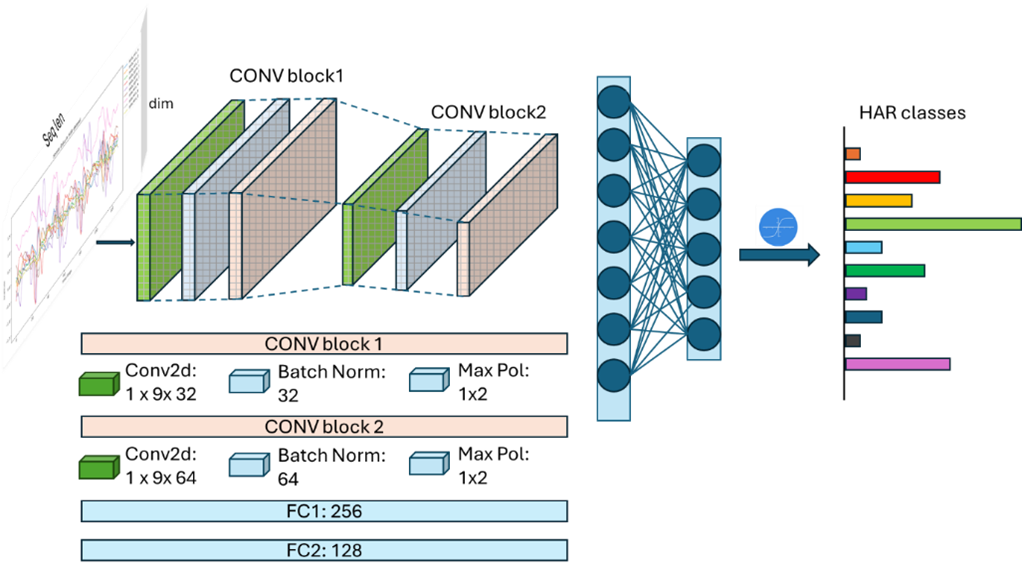}
\caption{Schematic diagram of the CNN-based model architecture. The model features two CONV blocks (with batch normalization and max pooling) and two FC layers designed for human activity recognition (HAR).}
\label{fig:HARCNN_Architecture}
\end{figure*}

The proposed architecture, illustrated in Figure \ref{fig:HARCNN_Architecture}, comprises a feature extraction backbone and a classification head. The backbone consists of two sequential convolutional blocks that progressively extract hierarchical representations, with each block incorporating a Conv2D layer ($1 \times 9$ filters), batch normalization, and max pooling ($1 \times 2$ window). The first block employs 32 filters to capture fundamental temporal features, while the second block increases to 64 filters to learn complex feature combinations. The classification head transforms these features through flattening, two fully-connected layers (256 and 128 neurons), and a softmax output layer to generate probability distributions over activity classes.

This design effectively captures temporal dependencies in sensor data while maintaining computational efficiency necessary for federated learning deployment. The progressive expansion of filter channels enables hierarchical feature learning from basic temporal patterns to activity-specific representations, while batch normalization throughout the architecture accelerates convergence, mitigates internal covariate shift, and provides implicit regularization against overfitting.

\subsection{The Complete Algorithm}
\label{method:algorithm}

The FedKDX training algorithm operates through a federated learning framework coordinated by a central server across multiple communication rounds, as systematically outlined in Algorithm \ref{alg:fedkdx}. The process begins with the server initializing a global lightweight student model and distributing it to all participating clients. Each client simultaneously initializes its own local teacher model, which remains private throughout the training process.

Each communication round begins with parallel client-side processing, where each client computes teacher and student loss functions as specified in Eq. (\ref{eq:fedkdx_teacher_loss}) and (\ref{eq:fedkdx_student_loss}). The client first updates its local teacher model using the computed teacher loss, then computes the student gradient. The client applies SVD-based compression to this gradient as described in Section \ref{method:gradient_compression}, with compression intensity controlled by an adaptive energy threshold from Eq. (\ref{eq:fedkdx_adaptiveT}). Only singular values meeting the threshold condition from Eq. (\ref{eq:fedkdx_threshold}) are transmitted to the server.

The server reconstructs individual client gradients and computes their weighted average to form the global gradient. The server then re-compresses this aggregated gradient using the same dynamic threshold from Eq. (\ref{eq:fedkdx_adaptiveT}) before broadcasting back to all clients.

Each communication round concludes with client-side synchronization, where clients receive the compressed global gradient, reconstruct it, and apply it to update their local student models. This process repeats iteratively across multiple communication rounds.

\begin{algorithm}[t]
\caption{Proposed FedKDX Framework}
\label{alg:fedkdx}
\begin{algorithmic}[1]
\State \textbf{Input:} Learning rates $\eta_T, \eta_S$; Client set $\mathcal{K}$
\State \textbf{Initialize:} Global student parameters $\mathbf{W}_S$, teacher parameters $\mathbf{W}_T^{(k)}$ for each client $k \in \mathcal{K}$.
\State Server distributes initial $\mathbf{W}_S$ to all clients.

\For{each communication round $t=1, 2, \dots$}
    \Statex \Comment{--- Client-side local computation ---}
    \For{each client $k \in \mathcal{K}$ \textbf{in parallel}}
        \State Compute losses $L_T^{(k)}$ and $L_S^{(k)}$ using Equations (\ref{eq:fedkdx_teacher_loss}) and (\ref{eq:fedkdx_student_loss}).
        \State Update local teacher model: $\mathbf{W}_T^{(k)} \leftarrow \mathbf{W}_T^{(k)} - \eta_T \nabla L_T^{(k)}(\mathbf{W}_T^{(k)})$.
        \State Compute student gradient $\nabla L_S^{(k)}(\mathbf{W}_S)$.
        \State Compress and send $\nabla L_S^{(k)}$ to the server based on threshold $\epsilon(\rho)$.
    \EndFor

    \Statex \Comment{--- Server-side aggregation and Client-side update ---}
    \State Server aggregates compressed gradients from all clients.
    \State All clients update their student model $\mathbf{W}_S$ according to Equation (\ref{eq:ws_update}).
\EndFor
\end{algorithmic}
\end{algorithm}

\section{Experiment}
\label{experiment}

\subsection{Datasets Overview and Evaluation Metrics}

\subsubsection{Dataset Overview}

\paragraph{Data Distribution Characteristics}

Independent and identically distributed (IID) data represents the ideal scenario where data samples across different clients follow the same underlying distribution, ensuring statistical homogeneity throughout the federated learning network. In contrast, non-IID data exhibits heterogeneous distributions across clients, reflecting real-world scenarios where different participants may have varying data characteristics, usage patterns, or demographic profiles. These distribution patterns significantly impact model evaluation, as IID conditions typically yield more stable and predictable performance metrics, while non-IID scenarios present greater challenges for model generalization and require more sophisticated evaluation approaches to assess robustness across diverse client environments.

\paragraph{SLEEP Dataset}

\begin{figure*}[t]%
\centering
\includegraphics[width=0.9\textwidth]{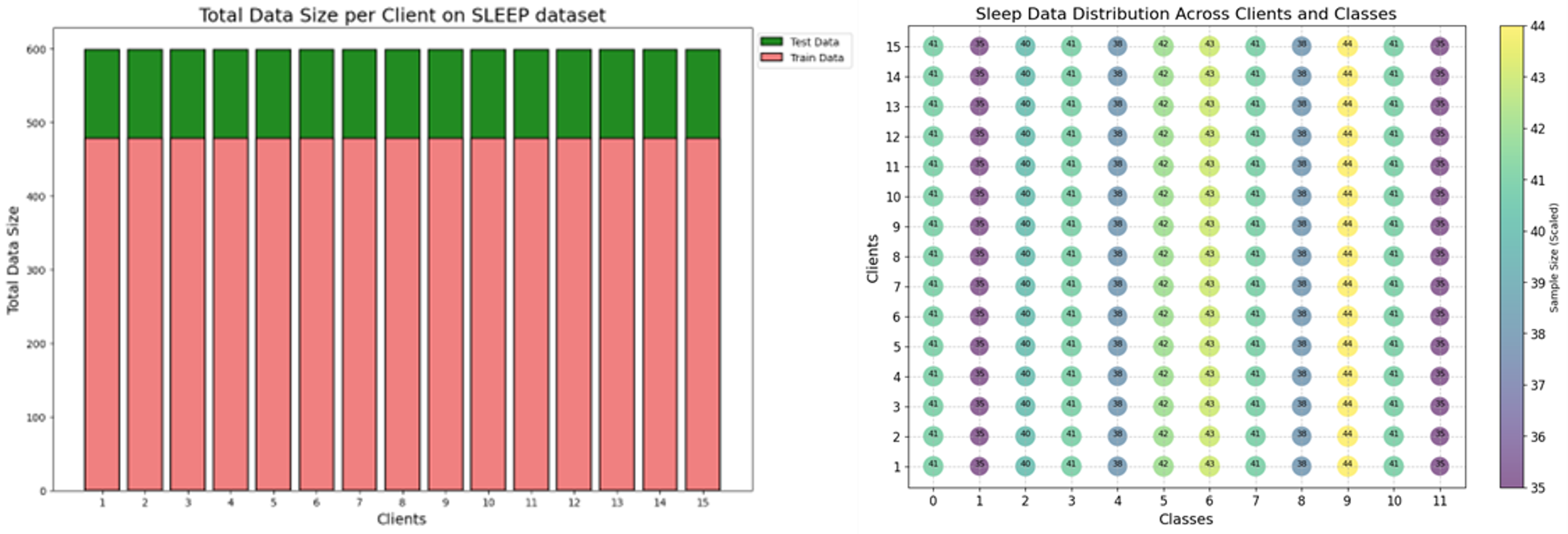}
\caption{Data distribution of the SLEEP dataset (16 participants, 12 postures). (Left) Total training and testing samples per client. (Right) Distribution of postures across participants, showing the balanced data allocation for sleep monitoring.}
\label{fig:SLEEP_Distribution}
\end{figure*}

The SLEEP dataset comprises sleep posture data collected from 16 adolescents with a mean age of $21 \pm 1$ years. This dataset captures 12 distinct static sleep postures recorded along three axes (X, Y, Z) using accelerometer sensors. The dataset structure includes 192 data files, with each file containing 3,000 samples representing one minute of posture data per position. The comprehensive nature of this dataset makes it particularly suitable for evaluating federated learning performance in healthcare monitoring applications, where data privacy and distributed processing are paramount concerns. Figure \ref{fig:SLEEP_Distribution} provides a visualization of the data distribution across clients, highlighting the balanced allocation of postures and participants.

\paragraph{UCI-HAR Dataset (Human Activity Recognition)}

\begin{figure*}[t]%
\centering
\includegraphics[width=0.9\textwidth]{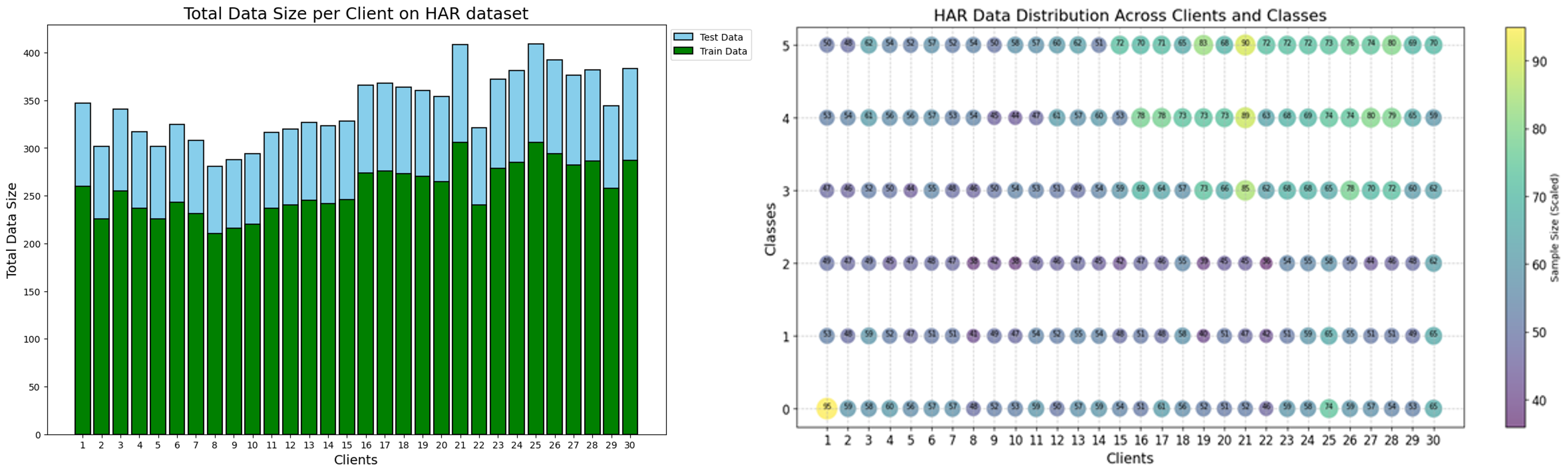}
\caption{Data distribution of the UCI-HAR dataset (30 subjects, 6 daily activities). (Left) Training and testing sample sizes per user. (Right) Class distribution across individuals, demonstrating the balanced nature of the controlled mobile sensing environment.}
\label{fig:HAR_Distribution}
\end{figure*}

The UCI-HAR dataset \cite{UCI_HAR} contains daily activity data from 30 participants aged between 19 and 48 years. The dataset encompasses six fundamental activities including walking, walking upstairs, walking downstairs, sitting, standing, and lying down. Data collection was performed using Samsung Galaxy S II smartphones worn at the waist, capturing both accelerometer and gyroscope measurements along three axes at a sampling frequency of 50\,Hz. This dataset serves as a standard benchmark for human activity recognition tasks and provides a robust foundation for evaluating federated learning algorithms in mobile sensing applications. A breakdown of the activity and participant distribution is presented in Figure \ref{fig:HAR_Distribution}.

\paragraph{PAMAP2 Dataset}

\begin{figure*}[t]%
\centering
\includegraphics[width=0.9\textwidth]{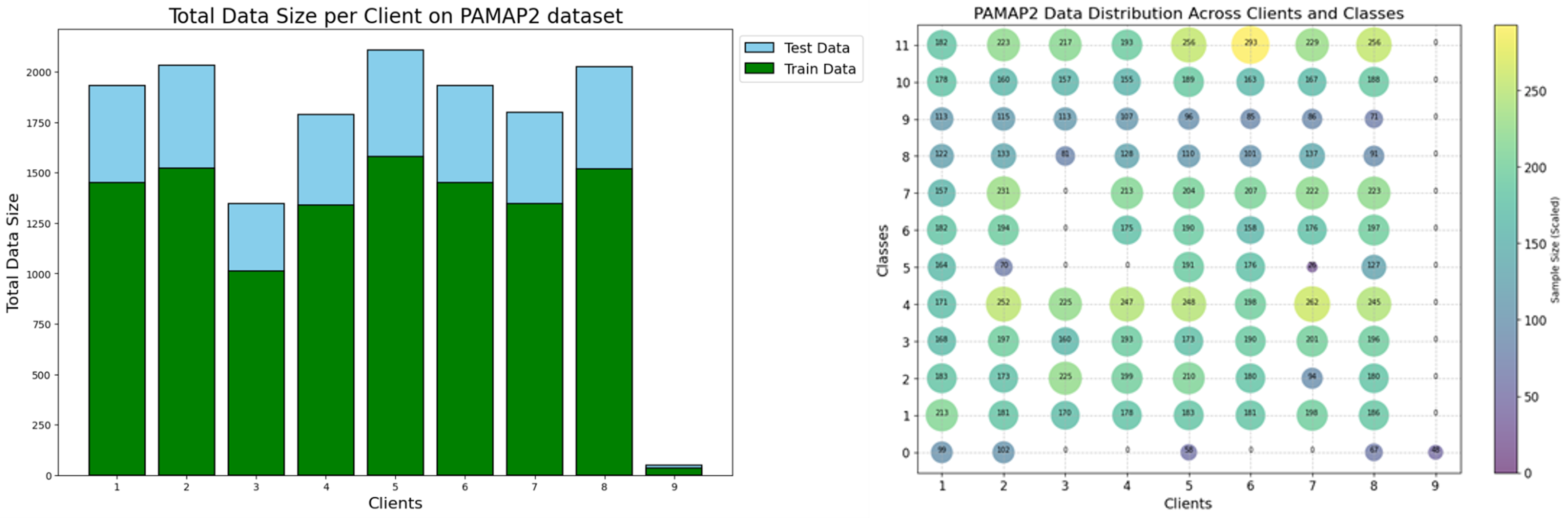}
\caption{Data distribution of the PAMAP2 dataset (9 subjects, 18 physical activities). (Left) Total samples per subject for training and testing. (Right) Activity coverage across users, reflecting the varying data density typical of multi-sensor activity recognition.}
\label{fig:PAMAP2_Distribution}
\end{figure*}

The PAMAP2 dataset \cite{PAMAP2} records 18 different physical activities from 9 subjects, each equipped with three Inertial Measurement Units (IMUs) positioned at the wrist, chest, and ankle. Each IMU captures acceleration, gyroscope, and magnetic field measurements, providing comprehensive motion data for activity recognition tasks. The dataset employs varying sampling frequencies, with IMU sensors operating at 100\,Hz and heart rate monitoring at approximately 9\,Hz. This multi-sensor approach offers rich temporal and spatial information for complex activity recognition scenarios. The heterogeneity in data volume and activity coverage across subjects is illustrated in Figure \ref{fig:PAMAP2_Distribution}.

For non-IID data simulation, a Dirichlet distribution with $\alpha=0.1$ is employed to create realistic heterogeneous data distributions across clients, mimicking real-world federated learning environments where data characteristics vary significantly between participants. 

\subsubsection{Evaluation Metrics}

The evaluation framework employs four primary metrics to assess model performance:

\begin{itemize}
    \item \textbf{Accuracy (\%)}: Measures the proportion of correctly classified samples out of total samples, providing an overall assessment of model performance across all activity classes.

    \item \textbf{F1-Score (\%)}: Represents the harmonic mean of precision and recall, offering a balanced evaluation metric that accounts for both false positives and false negatives, particularly valuable in imbalanced dataset scenarios.

    \item \textbf{Recall (\%)}: Quantifies the model's ability to correctly identify positive instances, measuring the proportion of actual positive cases that are correctly classified, crucial for applications where missing positive cases has significant consequences.

    \item \textbf{AUC (Area Under Curve)}: Evaluates the model's discriminative ability across different classification thresholds, providing a comprehensive measure of performance that is independent of specific threshold selections.
\end{itemize}

These metrics collectively provide a comprehensive evaluation framework that captures different aspects of model performance, from basic classification accuracy to more nuanced measures of precision, sensitivity, and discriminative power. In the context of activity recognition and behavioral assessment, these metrics help ensure that federated learning models maintain high performance standards while operating under distributed and potentially heterogeneous data conditions.

\subsection{Experiment Designs}

\subsubsection{Datasets pre-processing}

All datasets underwent a systematic preprocessing pipeline to ensure consistent feature extraction and enable fair architectural comparison. Raw inertial measurements from accelerometers and gyroscopes were filtered to reduce sensor noise. For the HAR and PAMAP2 datasets, we applied a median filter followed by a third-order low-pass Butterworth filter. Following the UCI-HAR reference protocol, acceleration signals were decomposed into body motion and gravitational components using a Butterworth low-pass filter with a 0.3 Hz cutoff frequency, thereby isolating dynamic movement patterns from static postural information. Continuous time-series data were then segmented into fixed-length windows using a sliding window approach. The UCI-HAR dataset employed a 2.56-second window (128 samples) with 50\% overlap between consecutive segments, capturing temporal dependencies while preserving transitional dynamics. Analogous segmentation parameters were adopted for the SLEEP and PAMAP2 datasets to maintain dimensional consistency with the CNN-based feature extractor.

Data distribution was organized by subject identifiers to authentically replicate federated learning conditions, ensuring each client represented a distinct user or user cohort. Local datasets were partitioned into 80\% training and 20\% testing subsets to evaluate client-level generalization. For independent and identically distributed (IID) benchmarks applied to SLEEP and HAR datasets, data were shuffled to balance class distributions across clients. To model statistical heterogeneity characteristic of healthcare applications (non-IID), we employed Dirichlet distribution partitioning with concentration parameter $\alpha=0.1$. This approach induces realistic inter-client imbalance wherein activity distributions vary substantially across participants, thereby testing global model robustness under heterogeneous data conditions.

\subsubsection{Method Comparision}

To provide a comprehensive benchmark for our proposed approach, we implemented and evaluated several state-of-the-art and foundational federated learning algorithms. These methods, detailed below, cover a range of strategies from foundational averaging to advanced personalization and knowledge distillation techniques:

\begin{itemize}
    \item \textbf{FedAvg (Federated Averaging) \cite{FedAvg}}: The foundational federated learning algorithm that trains a global model by aggregating locally computed model updates from distributed devices without requiring raw data transfer to a central server, thereby improving communication efficiency.

    \item \textbf{FedMAT (Federated Multi-task ATtention) \cite{FedMAT}}: A multi-task federated learning framework specifically designed for Human Activity Recognition (HAR) that enables learning of both shared and personalized features through attention mechanisms and centralized shared networks.

    \item \textbf{FedFomo (Federated First Order Model Optimization) \cite{FedFOMO}}: A personalized federated learning framework that allows each client to search and optimally combine models from other clients based on their utility for individual objectives, particularly effective with non-IID data and optimization for target distributions different from local data.

    \item \textbf{MOON (Model-Contrastive Federated Learning) \cite{MOON}}: A framework that addresses non-IID data heterogeneity by applying contrastive learning at the model level, aiming to reduce the distance between local model representations and the global model.

    \item \textbf{FedProx \cite{FedProx}}: A federated optimization framework that addresses system and statistical heterogeneity by constraining local updates to remain close to the initial global model while allowing devices to perform varying amounts of local work.

    \item \textbf{FedFTL (Federated Transfer Learning) \cite{FedFTL}}: A method that enables knowledge transfer between data domains when client datasets have minimal overlap, improving model accuracy while maintaining privacy preservation.

    \item \textbf{FedMTL (Federated Multi-Task Learning) \cite{FedMTL}}: A framework that leverages multi-task learning to address statistical challenges in federated learning, particularly with non-IID and imbalanced data, while providing system-aware optimization methods to handle practical issues such as communication costs and stragglers.

    \item \textbf{FedKD (Federated Knowledge Distillation) \cite{FedKD}}: A method that improves communication efficiency in federated learning through adaptive mutual knowledge distillation between large local teacher models and small global student models, combined with dynamic SVD-based gradient compression.
    
    \item \textbf{Proposed FedKDX}: A novel federated learning approach that combines Knowledge Distillation and Contrastive Learning with a new NKD loss function to improve model generalization and robustness, while utilizing dynamic SVD-based gradient compression to reduce communication costs.
\end{itemize}

\subsection{Implementation Details}

All experiments were conducted on NVIDIA A100 GPU hardware utilizing the PyTorch deep learning framework to ensure computational efficiency and reproducibility. The experimental environment provided sufficient computational resources to handle the complex federated learning scenarios and knowledge distillation processes required for comprehensive evaluation across multiple datasets and baseline methods.

The hyperparameter configuration was standardized across all experimental conditions to ensure a fair comparison between different federated learning approaches. The batch size was set to 32 for all training procedures, balancing memory efficiency with gradient estimation quality. The learning rate was configured at 0.01 to provide stable convergence across different model architectures and datasets. The number of participating clients in each federated round was set to match the number of participants in each dataset: 16 clients for the SLEEP dataset, 30 for UCI-HAR, and 9 for PAMAP2, ensuring a realistic simulation of user distribution in real-world monitoring scenarios. A fixed join ratio of 0.4 was applied uniformly across all methods, meaning 40\% of available clients were randomly selected to participate in each communication round. This value was chosen as it represents a practical trade-off between model performance and communication overhead, allowing sufficient aggregation stability while reducing bandwidth consumption and client-side resource demands in resource-constrained edge environments. All federated learning experiments were executed for 500 communication rounds, providing sufficient iterations for convergence analysis and performance evaluation.

Specific to the FedKDX implementation, the SVD compression threshold parameter ($\epsilon$) was set to 0.9, controlling the trade-off between communication efficiency and gradient fidelity during the dynamic compression process. The gamma parameter ($\gamma$) was configured at 0.9, regulating the contribution of the knowledge distillation components within the unified loss function. The tau parameter ($\tau$) was set to 0.8, governing the temperature scaling in both knowledge distillation and contrastive learning mechanisms to balance the probability distribution smoothness and sample pair discrimination. These parameter values were selected based on preliminary experiments to optimize the performance of the proposed method while maintaining computational feasibility across the federated learning scenarios.

\subsection{Ablation Study}

\subsubsection{Impact of Join Ratio}

In federated learning systems, the join ratio represents a critical hyperparameter that determines the number of clients participating in each training round. This parameter directly influences both model performance and communication costs, creating a fundamental trade-off between convergence quality and system efficiency. A higher join ratio typically leads to more stable model updates through increased data diversity but requires greater communication bandwidth and coordination overhead. Conversely, a lower join ratio reduces communication requirements but may result in less representative model updates and slower convergence rates.

To evaluate the impact of the join ratio on FedKDX performance, we conducted experiments on three benchmark human activity recognition datasets — HAR, SLEEP, and PAMAP2 — with join ratios ranging from 0.2 to 0.8. The evaluation focuses solely on accuracy, F1 score, and recall to measure how different participation rates influence the convergence behavior and final model performance. These results help identify the optimal trade-off between scalability and effectiveness, taking into account the knowledge distillation and contrastive learning components of the federated framework.

\subsubsection{Hybrid Model Architecture}
The study introduces a hybrid neural network architecture that combines Long Short-Term Memory (LSTM) networks with CNN to leverage the complementary strengths of both approaches for human activity recognition tasks. The LSTM component captures temporal dependencies and sequential patterns inherent in time-series sensor data, while the CNN layers with batch normalization extract robust spatial features and local patterns from the input signals. This hybrid design is particularly advantageous for human activity recognition scenarios where both temporal continuity and spatial feature extraction are crucial for accurate classification.

\begin{figure*}[t]%
\centering
\includegraphics[width=0.9\textwidth]{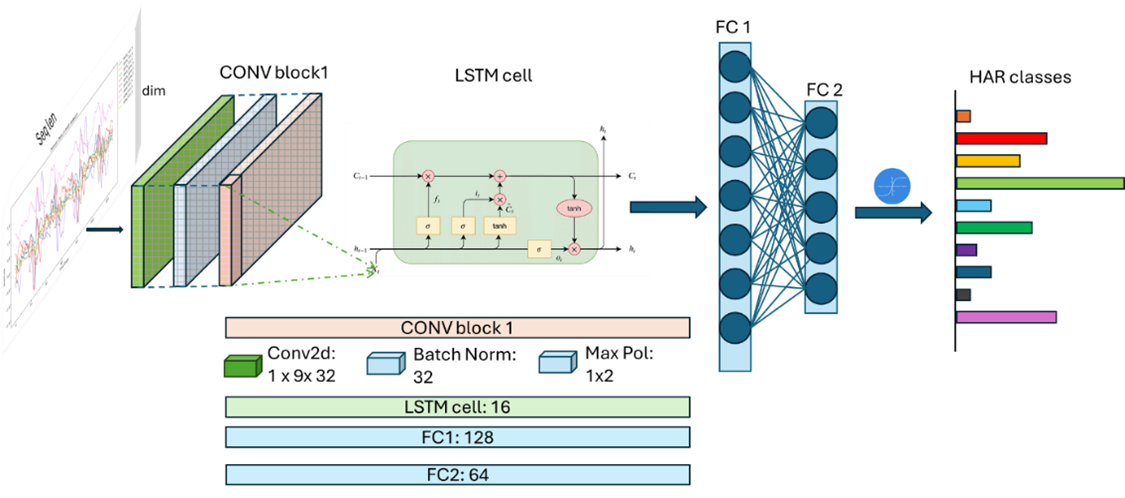}
\caption{Schematic of the hybrid CNN-LSTM model architecture. This design incorporates a CONV block for feature extraction followed by an LSTM cell and two FC layers for temporal activity classification.}
\label{fig:Hybrid_model}
\end{figure*}

The proposed hybrid architecture offers several anticipated advantages over pure CNN-based models, especially when processing time-series data typical in human activity recognition applications. The LSTM component provides memory capabilities that can model long-term dependencies across time steps, potentially improving recognition accuracy for activities with extended duration patterns. The CNN layers contribute efficient feature extraction and translation invariance, while batch normalization enhances training stability and convergence speed. The detailed architecture specifications and layer configurations are illustrated in Figure \ref{fig:Hybrid_model}, demonstrating how the temporal and spatial processing components are integrated within the federated learning framework.

\subsubsection{Impact of NKD and Contrastive learning Loss Functions}

To evaluate the individual contributions of the Negative Knowledge Distillation and Contrastive Learning components within FedKDX, systematic ablation experiments were conducted by selectively removing these loss function components. These experiments isolate the specific impact of each proposed enhancement on overall model performance and convergence characteristics. The NKD loss function is designed to prevent the student model from learning incorrect patterns by explicitly penalizing high confidence predictions on negative examples, while the CTL component encourages the model to learn discriminative representations by contrasting positive and negative sample pairs.

The ablation study examines model performance across multiple metrics when training with and without these components, providing quantitative evidence of their respective contributions to the final system performance. By systematically removing the NKD and CTL loss terms, the experiments reveal how each component affects model generalization, robustness to data heterogeneity, and overall federated learning efficiency. These results demonstrate the necessity and effectiveness of each proposed enhancement within the comprehensive FedKDX framework, supporting the theoretical motivations for their inclusion in the unified loss function design.

\section{Results}
\label{results}

\subsection{Overall Performance Comparison}

Table~\ref{tab:harcnn_results} presents the comparative results of seven different federated learning methods across three datasets using CNN-based architectures, evaluated on diverse metrics including Accuracy, AUC, F1-score, and Recall on the test datasets. The results demonstrate that FedKDX exhibits superior performance compared to all baseline methods across all datasets, revealing significant insights about the effectiveness of negative knowledge distillation in addressing federated learning challenges under both IID and non-IID data conditions.

FedKDX achieves remarkable accuracy scores of 94.54\% on the HAR dataset, 98.06\% on SLEEP, and 89.85\% on PAMAP2, surpassing the second-best methods by margins of 0.73\%, 0.50\%, and 2.53\% respectively. The most significant improvements are observed in the AUC metric, where FedKDX achieves near-perfect scores of 0.9944, 0.9978, and 0.9826 across the three datasets. This indicates that FedKDX not only achieves high accuracy but also develops exceptionally robust decision boundaries that maintain performance across varying classification thresholds.

While FedMTL demonstrates comparable performance to FedKDX under standard conditions, the performance gap becomes significantly larger under statistical heterogeneity conditions, particularly when considering AUC, F1-score, and Recall metrics. FedKD, despite employing similar knowledge distillation principles, shows substantial differences compared to FedKDX, highlighting the critical importance of the comprehensive knowledge transfer approach proposed in the proposed method.

The superior AUC scores achieved by FedKDX across all datasets indicate enhanced discriminative capability, suggesting that the negative knowledge distillation mechanism successfully captures nuanced class relationships that improve model robustness. This discriminative power is particularly valuable in applications where precise classification boundaries are essential for reliable decision-making.

\begin{table}[t]
\centering
\caption{Performance comparison of federated learning methods using a CNN architecture. Baseline methods are presented in chronological order. Metrics include accuracy, AUC, F1-score, and recall across three healthcare activity recognition tasks. The symbol $\uparrow$ indicates that higher values are better.}
\label{tab:harcnn_results}
\resizebox{\textwidth}{!}{%
\begin{tabular}{@{}ccccccccccccc@{}}
\toprule
\multirow{2}{*}{\textbf{Method}} & \multicolumn{4}{c}{\textbf{HAR}} & \multicolumn{4}{c}{\textbf{SLEEP}} & \multicolumn{4}{c}{\textbf{PAMAP2}} \\
\cmidrule(lr){2-5} \cmidrule(lr){6-9} \cmidrule(lr){10-13}
 & Acc $\uparrow$ & AUC $\uparrow$ & F1 $\uparrow$ & Recall $\uparrow$ & Acc $\uparrow$ & AUC $\uparrow$ & F1 $\uparrow$ & Recall $\uparrow$ & Acc $\uparrow$ & AUC $\uparrow$ & F1 $\uparrow$ & Recall $\uparrow$ \\
\midrule
FedAvg \cite{FedAvg}     & 85.06\% & 96.10\% & 82.46\% & 84.68\% & 81.17\% & 96.00\% & 79.77\% & 82.67\% & 72.43\% & 91.67\% & 59.47\% & 60.70\% \\
FedMTL \cite{FedMTL}     & 93.81\% & 96.17\% & 92.80\% & 92.93\% & 97.56\% & 94.04\% & 97.34\% & 97.71\% & 87.32\% & 93.40\% & 86.72\% & 86.77\% \\
FedProx \cite{FedProx}   & 85.14\% & 96.20\% & 83.09\% & 85.09\% & 81.22\% & 95.98\% & 79.83\% & 82.71\% & 73.12\% & 91.63\% & 60.41\% & 61.52\% \\
FedFomo \cite{FedFOMO}   & 92.88\% & 95.95\% & 91.69\% & 92.01\% & 97.00\% & 94.07\% & 96.55\% & 97.09\% & 84.87\% & 93.14\% & 73.14\% & 73.50\% \\
MOON \cite{MOON}         & 85.06\% & 96.10\% & 82.46\% & 84.68\% & 81.17\% & 95.97\% & 79.77\% & 82.67\% & 73.15\% & 91.45\% & 60.30\% & 61.53\% \\
FedKD \cite{FedKD}       & 74.88\% & 96.80\% & 69.55\% & 73.01\% & 96.17\% & 99.15\% & 95.59\% & 96.29\% & 85.83\% & 97.47\% & 86.60\% & 84.78\% \\
FedKDX (Ours)            & \textbf{94.54\%} & \textbf{99.44\%} & \textbf{93.73\%} & \textbf{93.83\%} & \textbf{98.06\%} & \textbf{99.78\%} & \textbf{97.92\%} & \textbf{98.25\%} & \textbf{89.85\%} & \textbf{98.26\%} & \textbf{89.37\%} & \textbf{89.32\%} \\
\bottomrule
\end{tabular}%
}
\end{table}

\subsection{Convergence Analysis}

\begin{figure*}[t]%
\centering
\includegraphics[width=1.0\textwidth]{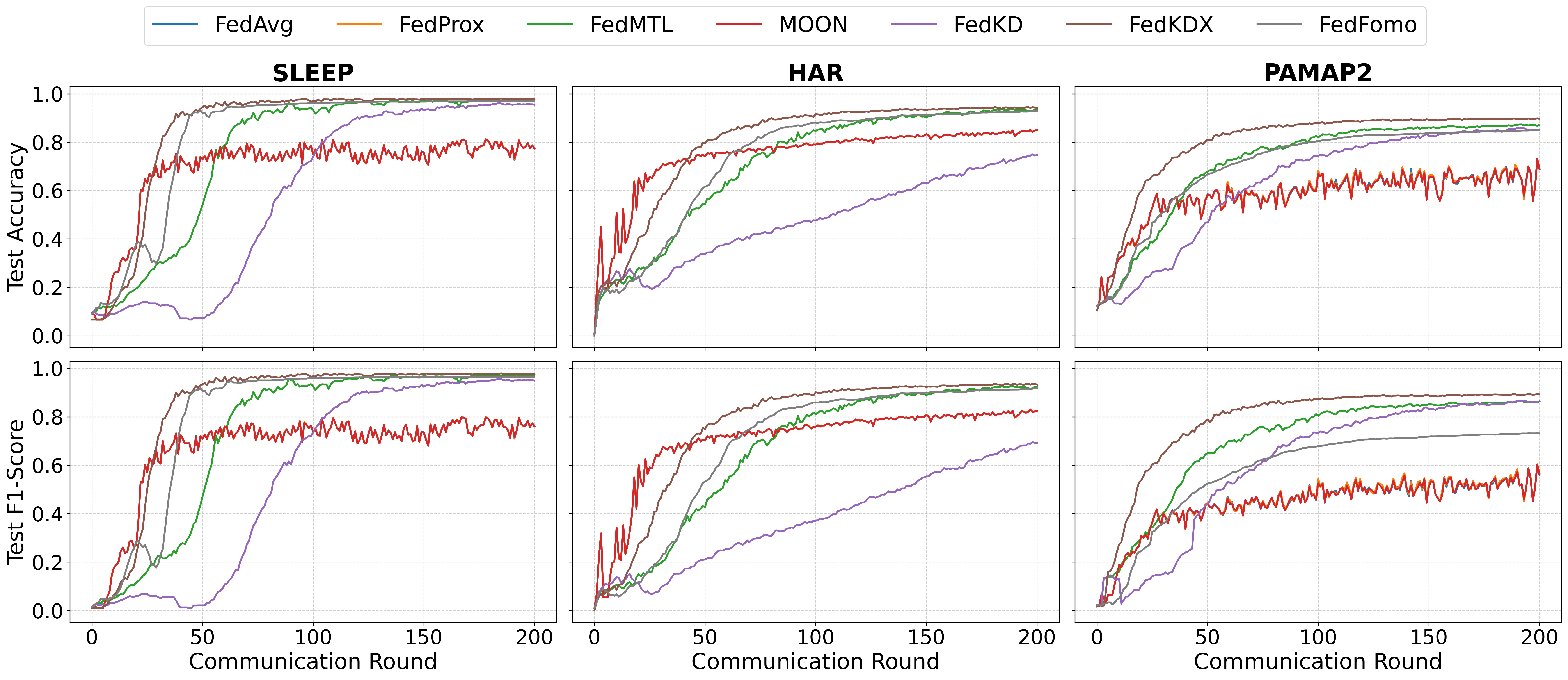}
\caption{Training convergence comparison across three datasets. Accuracy and F1-score trajectories are shown over 200 communication rounds for the SLEEP, HAR, and PAMAP2 datasets.}
\label{fig:training_curves}
\end{figure*}

Figure~\ref{fig:training_curves} illustrates the training accuracy and F1-score curves for all seven federated learning methods across the three datasets. The results demonstrate that FedKDX and FedFomo consistently achieve the highest performance on all three datasets for both evaluation metrics, exhibiting rapid convergence while maintaining high and stable accuracy and F1-scores. This superior performance is particularly pronounced on the SLEEP dataset, where both methods demonstrate clear advantages over competing approaches.

FedMTL, while achieving competitive results comparable to FedKDX and maintaining good convergence properties, exhibits significantly slower convergence rates compared to both FedKDX and FedFomo. This difference in convergence speed could impact practical deployment scenarios where communication efficiency is critical.

MOON demonstrates the fastest convergence in the initial communication rounds but fails to achieve high final performance levels. Additionally, MOON exhibits the largest fluctuations in its convergence curves, indicating training instability that could compromise the reliability of the federated learning process. This behavior suggests that while rapid initial convergence may appear advantageous, sustained performance improvements require more sophisticated knowledge transfer mechanisms.

\subsection{Ablation Studies}

\subsubsection{Analysis of losses component}
\label{subsubsec:loss_analysis}

The ablation study examining individual contributions of Negative Knowledge Distillation and Contrastive Learning components provides crucial insights into the framework design, as presented in Table~\ref{tab:ablation_study}. While the individual improvements from adding NKD and CTL components may appear modest in absolute terms, they represent noticeable and consistent enhancements across all datasets and metrics.

The progressive addition of components demonstrates a synergistic effect, where the combination of NKD and CTL yields greater benefits than the sum of individual contributions. This pattern suggests that structural alignment through contrastive learning enhances the effectiveness of negative knowledge distillation by providing more stable feature representations, while negative knowledge distillation sharpens decision boundaries to complement the structural improvements.

The consistency of improvements across different datasets, despite varying data characteristics and distribution patterns, validates the universal applicability of the proposed mechanisms. Even modest improvements in federated learning scenarios can translate to significant practical benefits when considering the challenges of distributed training and privacy preservation.

\begin{table}[t]
\centering
\caption{Ablation study of the FedKDX framework. Impact of negative knowledge distillation (NKD) and contrastive learning (CTL) components on model performance.}
\label{tab:ablation_study}
\begin{tabular}{@{}llcccc@{}}
\toprule
\textbf{Dataset} & \textbf{Components} & \textbf{Acc $\uparrow$} & \textbf{AUC $\uparrow$} & \textbf{F1 $\uparrow$} & \textbf{Recall $\uparrow$} \\
\midrule
\multirow{3}{*}{HAR} 
 & Base & 93.96\% & \textbf{99.38\%} & 93.02\% & 93.13\% \\
 & Base + NKD & 93.96\% & \textbf{99.38\%} & 93.07\% & 93.12\% \\
 & Base + CT + NKD & \textbf{94.08\%} & 99.30\% & \textbf{93.15\%} & \textbf{93.27\%} \\
\midrule
\multirow{3}{*}{SLEEP} 
 & Base & 97.94\% & 99.60\% & 97.76\% & 98.06\% \\
 & Base + NKD & 98.00\% & \textbf{99.69\%} & 97.86\% & 98.17\% \\
 & Base + CT + NKD & \textbf{98.11\%} & 99.67\% & \textbf{97.98\%} & \textbf{98.21\%} \\
\midrule
\multirow{3}{*}{PAMAP2} 
 & Base & 93.85\% & 99.03\% & 93.52\% & 93.34\% \\
 & Base + NKD & 93.74\% & 98.83\% & 93.63\% & 93.28\% \\
 & Base + CT + NKD & \textbf{94.03\%} & \textbf{99.18\%} & \textbf{93.68\%} & \textbf{93.50\%} \\
\bottomrule
\end{tabular}
\end{table}

\subsubsection{Analysis of different Join Ratio}
\label{subsubsec:join_ratio_analysis}

Table~\ref{tab:join_ratio_ablation_reordered} presents a comprehensive analysis of how client participation rates affect FedKDX performance across different datasets. The join ratio experiments reveal important insights about the scalability and robustness of the proposed framework under varying levels of client participation.

For the HAR dataset, performance improvements are substantial when increasing the join ratio from 0.2 to 0.4, with accuracy jumping from 78.23\% to 90.80\%. This steep improvement curve suggests that FedKDX benefits significantly from increased client diversity in the early stages. However, the marginal improvements become smaller at higher join ratios, with accuracy increasing from 93.16\% at 0.7 to 93.50\% at 0.8, indicating diminishing returns. The training time increases linearly with join ratio, from 456.25 seconds at 0.2 participation to 1200.26 seconds at 0.8 participation, highlighting the communication cost trade-offs.

The SLEEP dataset demonstrates more graceful performance scaling, achieving 96.06\% accuracy even at 0.2 join ratio and reaching 97.72\% at 0.8 participation. This behavior suggests that FedKDX's knowledge distillation mechanisms provide effective information transfer even with limited client participation, making it particularly suitable for scenarios where client availability is constrained.

The PAMAP2 dataset, with its non-IID characteristics, shows the most dramatic improvements with increased participation. At 0.2 join ratio, accuracy is only 72.56\%, but it increases substantially to 84.87\% at 0.3 participation and continues improving to 89.48\% at 0.8 participation. This pattern indicates that non-IID data requires higher client participation to achieve optimal performance, as increased diversity helps counteract the negative effects of statistical heterogeneity.

\begin{table}[t]
\centering
\caption{Effect of client participation rate on FedKDX performance. The table presents evaluation metrics and corresponding training time requirements across different join ratios.}
\label{tab:join_ratio_ablation_reordered}
\begin{tabular}{@{}lcccccc@{}} 
\toprule
\textbf{Dataset} & \textbf{Join Ratio} & \textbf{Acc $\uparrow$} & \textbf{AUC $\uparrow$} & \textbf{F1 $\uparrow$} & \textbf{Recall $\uparrow$} & \textbf{Training Time (s)} $\downarrow$\\
\midrule
\multirow{7}{*}{HAR} 
 & 0.2 & 78.23\% & 96.72\% & 74.05\% & 77.61\% & 456.25  \\
 & 0.3 & 86.67\% & 97.55\% & 84.69\% & 85.96\% & 549.04  \\
 & 0.4 & 90.80\% & 98.39\% & 89.83\% & 90.42\% & 629.86  \\
 & 0.5 & 91.65\% & 98.66\% & 90.77\% & 91.37\% & 727.18  \\
 & 0.6 & 92.57\% & 98.82\% & 91.68\% & 92.27\% & 853.46  \\
 & 0.7 & 93.16\% & 99.10\% & 92.58\% & 92.89\% & 937.97  \\
 & 0.8 & \textbf{93.50\%} & \textbf{99.27\%} & \textbf{92.91\%} & \textbf{93.23\%} & 1200.26 \\
\midrule
\multirow{7}{*}{SLEEP} 
 & 0.2 & 96.06\% & 97.34\% & 95.63\% & 96.40\% & 743.72  \\
 & 0.3 & 97.00\% & 98.31\% & 96.59\% & 97.13\% & 881.67  \\
 & 0.4 & 97.44\% & 99.54\% & 97.21\% & 97.59\% & 1007.07 \\
 & 0.5 & 97.56\% & 99.57\% & 97.33\% & 97.71\% & 1307.82 \\
 & 0.6 & 97.67\% & \textbf{99.66\%} & 97.47\% & 97.83\% & 1603.10 \\
 & 0.7 & 97.67\% & \textbf{99.66\%} & 97.47\% & 97.84\% & 1752.70 \\
 & 0.8 & \textbf{97.72\%} & \textbf{99.66\%} & \textbf{97.53\%} & \textbf{97.89\%} & 2040.29 \\
\midrule
\multirow{7}{*}{PAMAP2} 
 & 0.2 & 72.56\% & 93.69\% & 68.38\% & 70.27\% & 1380.99 \\
 & 0.3 & 84.87\% & 97.26\% & 83.46\% & 83.22\% & 1782.88 \\
 & 0.4 & 87.72\% & 97.90\% & 87.02\% & 86.96\% & 2010.06 \\
 & 0.5 & 88.81\% & 98.08\% & 88.16\% & 87.98\% & 2605.66 \\
 & 0.6 & 89.00\% & 98.07\% & 88.44\% & 88.28\% & 2994.18 \\
 & 0.7 & 89.40\% & 98.01\% & 88.88\% & 88.86\% & 3367.44 \\
 & 0.8 & \textbf{89.48\%} & \textbf{98.17\%} & \textbf{88.97\%} & \textbf{88.91\%} & 3527.27 \\
\bottomrule
\end{tabular}
\end{table}

\subsubsection{Analysis of Model Architectures}

The evaluation of the hybrid CNN-LSTM architecture shows that FedKDX effectively adapts to different neural network designs while preserving its performance advantages. This hybrid model excels on datasets with strong temporal dependencies, such as HAR and PAMAP2, where capturing sequential patterns is crucial, while CNN-based models remain superior for data emphasizing static or spatial features, demonstrating the flexibility of FedKDX's knowledge distillation mechanism. Thanks to this adaptability, practitioners can select architectures tailored to their domain-specific requirements without losing the benefits of negative knowledge distillation. The hybrid approach, however, comes with trade-offs: the LSTM's sequential processing and extra parameters increase computational complexity and memory usage, which can hinder training efficiency in resource-constrained environments. Even so, the performance gains on temporal data often justify these costs, especially in applications where precise temporal modeling is critical.

The results presented in Table~\ref{tab:hybrid_bn_results} show that FedKDX continues to demonstrate superior performance, achieving the highest scores on nearly all datasets with the hybrid architecture, with the notable exception of the HAR dataset where it marginally underperforms compared to FedMTL. The remaining baseline methods also show significant improvements with the hybrid model structure, indicating that while FedKDX may not exhibit perfect scalability with increasing model complexity, it still demonstrates good scalability properties with substantial performance enhancements across different architectural configurations.

\begin{table}[t]
\centering
\caption{Performance evaluation of federated learning methods with a hybrid CNN-LSTM architecture. Baseline models are listed chronologically to illustrate the progression in healthcare activity recognition tasks.}
\label{tab:hybrid_bn_results}
\resizebox{\textwidth}{!}{%
\begin{tabular}{@{}ccccccccccccc@{}}
\toprule
\multirow{2}{*}{\textbf{Method}} & \multicolumn{4}{c}{\textbf{HAR}} & \multicolumn{4}{c}{\textbf{SLEEP}} & \multicolumn{4}{c}{\textbf{PAMAP2}} \\
\cmidrule(lr){2-5} \cmidrule(lr){6-9} \cmidrule(lr){10-13}
 & Acc $\uparrow$ & AUC $\uparrow$ & F1 $\uparrow$ & Recall $\uparrow$ & Acc $\uparrow$ & AUC $\uparrow$ & F1 $\uparrow$ & Recall $\uparrow$ & Acc $\uparrow$ & AUC $\uparrow$ & F1 $\uparrow$ & Recall $\uparrow$ \\
\midrule
FedAvg \cite{FedAvg}     & 93.00\% & 98.49\% & 92.36\% & 92.85\% & 83.11\% & 96.35\% & 80.99\% & 84.23\% & 86.49\% & 94.85\% & 82.23\% & 83.01\% \\
FedMTL \cite{FedMTL}     & \textbf{94.54\%} & 97.57\% & \textbf{93.83\%} & \textbf{94.16\%} & 97.83\% & 98.06\% & 97.77\% & 98.05\% & 96.14\% & 96.44\% & 95.98\% & 95.90\% \\
FedProx \cite{FedProx}   & 92.96\% & 98.49\% & 92.37\% & 92.82\% & 83.39\% & 96.38\% & 81.25\% & 84.42\% & 87.29\% & 94.02\% & 87.91\% & 88.30\% \\
FedFomo \cite{FedFOMO}   & 89.67\% & 97.12\% & 87.93\% & 88.70\% & 97.67\% & 97.59\% & 97.63\% & 97.94\% & 94.17\% & 95.55\% & 82.89\% & 82.90\% \\
MOON \cite{MOON}         & 93.11\% & 98.48\% & 92.48\% & 92.98\% & 83.61\% & 96.34\% & 81.57\% & 84.67\% & 87.37\% & 94.47\% & 77.69\% & 77.53\% \\
FedKD \cite{FedKD}       & 92.65\% & 98.82\% & 91.71\% & 91.73\% & 97.33\% & \textbf{99.19\%} & 97.09\% & 97.43\% & 95.10\% & \textbf{98.71\%} & 95.54\% & 94.84\% \\
FedKDX (Ours)            & 94.50\% & \textbf{99.15\%} & 93.56\% & 93.91\% & \textbf{98.72\%} & 98.25\% & \textbf{98.69\%} & \textbf{98.87\%} & \textbf{96.35\%} & 96.16\% & \textbf{96.07\%} & \textbf{96.03\%} \\
\bottomrule
\end{tabular}%
}
\end{table}

\section{Discussion}
\label{discussion}

\subsection{Effectiveness of Negative Knowledge Distillation}

The experimental results demonstrate that Negative Knowledge Distillation provides superior performance compared to traditional positive knowledge transfer approaches. The NKD mechanism captures comprehensive class relationships by explicitly modeling non-target information, proving particularly effective in healthcare applications where understanding boundaries between different activities is crucial. The consistent performance improvements across all datasets, especially the enhanced AUC scores, indicate that FedKDX develops more discriminative decision boundaries essential for reliable medical diagnosis systems.

\subsection{Scalability and Communication Efficiency}

The join ratio analysis reveals dataset-dependent scalability characteristics. FedKDX maintains competitive performance with limited client participation for homogeneous data distributions but requires higher participation rates for non-IID scenarios. The framework achieves significant communication efficiency advantages through compressed gradient exchange while preserving privacy guarantees required by healthcare regulations. The dynamic threshold strategy effectively balances communication costs with model accuracy throughout training.

\subsection{SVD Non-Convergence Issue in FedKDX}

The SVD non-convergence phenomenon in FedKDX arises from the fundamental evolution of gradient matrix spectral properties during federated training under non-IID conditions. SVD-based compression assumes gradient matrices maintain stable low-rank structures, enabling approximation through dominant singular components. However, empirical evidence reveals that singular value energy becomes increasingly dispersed as training progresses, indicating more components contribute meaningfully to reconstruction and reflecting increased effective rank. This dispersion directly contradicts the low-rank assumptions essential for efficient SVD compression.

Early training phases produce large-scale, structured gradient updates that align with principal parameter directions, yielding well-conditioned decompositions with few dominant singular values. As optimization approaches convergence, updates become fine-grained and heterogeneous across clients, particularly under data heterogeneity. These later-stage gradients resist low-rank approximation, requiring substantially more singular components for adequate reconstruction fidelity. This necessity eliminates compression benefits while increasing numerical sensitivity and computational complexity.

Training instability compounds these challenges when conflicting client updates or poorly conditioned loss landscapes prevent smooth convergence. Gradient matrices inherit erratic characteristics that manifest as fluctuating singular vectors and values across iterations, preventing stable decomposition. The algorithm may fail numerically when matrices lack clear spectral gaps or exhibit near-degenerate singular values, reflecting fundamental incompatibility between static SVD methods and dynamic distributed optimization rather than mere computational limitations.

The SVD non-convergence therefore represents deeper theoretical tensions between low-rank compression assumptions and federated learning dynamics. As gradient matrices evolve from structured, low-rank forms to spectrally dispersed configurations, the foundational requirements for stable SVD operation are violated, compromising both compression fidelity and training stability.

\subsection{Limitation and Future Work}
Despite its strong performance, FedKDX faces limitations that warrant further study. SVD-based gradient compression exhibits instability in later training stages, suggesting the need for adaptive or more robust compression methods that account for evolving gradient spectra. Additionally, while improved under heterogeneous data, performance can degrade under extreme distributional shifts, indicating room for enhanced personalization strategies. Scalability to large client networks also requires evaluation in real-world, large-scale settings. Finally, current validation is limited to activity and posture recognition; extending evaluation to medical imaging, EHRs, and multi-modal data would strengthen evidence of its broad applicability in healthcare AI.

\section{Conclusion}
\label{conclusion}

This paper introduces FedKDX, a federated learning framework that integrates Negative Knowledge Distillation, contrastive learning, and dynamic gradient compression to address critical limitations in existing approaches. The framework achieves superior performance across healthcare datasets, with accuracy improvements up to 2.53\% over state-of-the-art methods while maintaining enhanced AUC scores and communication efficiency.

The key contributions include the development of comprehensive knowledge transfer mechanisms that capture both positive and negative class relationships, successful integration of multiple learning paradigms within a unified architecture, and extensive empirical validation demonstrating improved robustness under both IID and non-IID conditions. The framework's architectural flexibility and privacy-preserving capabilities make it particularly suitable for real-world healthcare AI deployments.

Future work should address SVD compression instability, investigate scalability to larger client populations, and expand evaluation to diverse healthcare applications. The principles developed in FedKDX extend beyond healthcare to other privacy-sensitive domains, representing an important advancement toward realizing the full potential of federated learning in collaborative AI systems.

\printbibliography

\end{document}